\documentclass[letterpaper, 10 pt, conference]{ieeeconf}  

\IEEEoverridecommandlockouts                              

\usepackage{cite}
\usepackage{booktabs}
\usepackage{graphicx} 
\usepackage{amsmath} 
\usepackage{amssymb}  
\usepackage{pifont}
\usepackage{url}
\usepackage{subcaption}

\title{\LARGE \bf
Autoregressive End-to-End Planning with Time-Invariant Spatial Alignment and Multi-Objective Policy Refinement
}

\author{
    Jianbo Zhao$^{1,2,*}$, Taiyu Ban$^{1,2,*}$, Xiangjie Li$^{2,*}$, Xingtai Gui$^{2}$, Hangning Zhou$^{2}$, \\Lei Liu$^{1, \dagger}$, Hongwei Zhao$^1$, Bin Li$^1$,
    \thanks{$^*$ These authors contributed equally to this work.}
    \thanks{$^\dagger$ Corresponding authors: Lei Liu (liulei13@ustc.edu.cn).}
    \thanks{$^1$ Jianbo Zhao (zjb123@mail.ustc.edu.cn), Taiyu Ban, Hongwei Zhao, Lei Liu, and Bin Li are with University of Science and Technology of China, 96 Jinzhai Rd, Hefei 230026, China.}
    \thanks{$^2$ Jianbo Zhao, Taiyu Ban, Xiangjie Li, Xingtai Gui, and Hangning Zhou are with Mach Drive, Suzhou Rd 3, Beijing 100080, China.}
}

\begin{document}

\maketitle
\thispagestyle{empty}
\pagestyle{empty}

\begin{abstract}

The inherent sequential modeling capabilities of autoregressive models make them a formidable baseline for end-to-end planning in autonomous driving. Nevertheless, their performance is constrained by a spatio-temporal misalignment, as the planner must condition future actions on past sensory data. This creates an inconsistent worldview, limiting the upper bound of performance for an otherwise powerful approach.
To address this, we propose a Time-Invariant Spatial Alignment (TISA) module that learns to project initial environmental features into a consistent ego-centric frame for each future time step, effectively correcting the agent's worldview without explicit future scene prediction. In addition, we employ a kinematic action prediction head (i.e., acceleration and yaw rate) to ensure physically feasible trajectories. Finally, we introduce a multi-objective post-training stage using Direct Preference Optimization (DPO) to move beyond pure imitation. Our approach provides targeted feedback on specific driving behaviors, offering a more fine-grained learning signal than the single, overall objective used in standard DPO. Our model achieves a state-of-the-art 89.8 PDMS on the NAVSIM dataset among autoregressive models. The video document is available at \url{https://tisa-dpo-e2e.github.io/}.

\end{abstract}

\section{INTRODUCTION}

End-to-end autonomous driving aims to learn a driving policy directly from raw sensor inputs \cite{hu2023planning}, bypassing the cascaded perception and planning modules of traditional frameworks and their associated cumulative errors \cite{seff2023motionlm}. Prominent end-to-end approaches include trajectory regression \cite{jiang2023vad}, autoregressive models \cite{chen2024vadv2}, and diffusion models \cite{liao2025diffusiondrive}. Regression methods are known to suffer from the mode averaging issue \cite{zhao2024kigras}. The other methods are generative approaches and can capture multi-modal driving behaviors. Among them, autoregressive models offer unique flexibility by explicitly modeling the action probability distribution, facilitating straightforward integration with techniques like reinforcement learning \cite{gao2025rad}.

\begin{figure}[!t]
    \centering
    \includegraphics[width=0.98\linewidth]{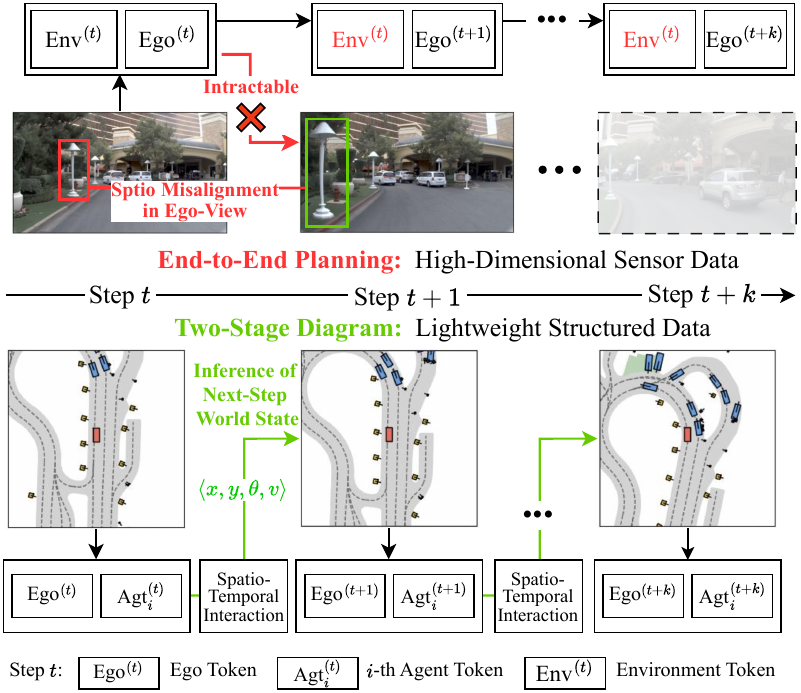}
    \caption{Spatio-temporal misalignment in autoregressive planning. \textbf{(Below)} A two-stage model maintains alignment by inferring future structured world states ($S_{t+k}$). \textbf{(Top)} An end-to-end model reuses a stale environmental view from the present ($t$) to plan for a future step ($t+k$), creating a critical misalignment between the agent's prospective state and its perceived world.}
    \label{fig:issue}
\end{figure}

However, autoregressive end-to-end planning presents a unique challenge: maintaining a consistent worldview for future steps. In modular, two-stage pipelines, this is often resolved by predicting future environmental states, which is a tractable task when operating on lightweight, structured data \cite{wu2024smart}. This allows the planner to act within a consistently updated, inferred world \cite{zhao2024kigras}.
This strategy, however, is intractable in the end-to-end paradigm where predicting future high-dimensional sensor data is computationally infeasible. Consequently, end-to-end models typically base their entire plan on merely past snapshots of the world \cite{chen2024vadv2}. This creates a critical \textbf{spatio-temporal misalignment}: the agent plans for a future step $t+k$ while perceiving the world from a stale viewpoint at time $t$. This issue is illustrated in Fig. \ref{fig:issue}.

To resolve this spatio-temporal misalignment, we propose a novel alignment mechanism in the latent space. Our approach is motivated by the observation that an ego-view update, which consists of rotation and translation, can be modeled as a time-invariant spatial transformation. We hypothesize that this transformation can be learned and applied within a latent space to align environmental features with future ego states. To realize this, we introduce the Time-Invariant Spatial Alignment (TISA) module. For each future time step, TISA processes a query that fuses static environmental features with the ego-vehicle's prospective state. In response, the module outputs a spatially-aligned environmental context, ensuring that the model's subsequent action prediction is always conditioned on a geometrically consistent and relevant worldview.

Moreover, we introduce two additional components to enhance the realism and quality of the learned driving policy.
First, instead of predicting spatial waypoints, our model outputs discretized kinematic actions (i.e., acceleration and yaw rate). This approach not only creates a more compact action space but also ensures that the generated trajectories are inherently physically feasible. To move beyond the limitations of simple imitation learning, we introduce a multi-objective post-training stage using DPO \cite{rafailov2023direct}. We construct a preference dataset by generating diverse trajectories and labeling them using multiple safety-focused metrics\footnote{Our fine-tuning prioritizes safety, as the model's kinematic action space already produces trajectories that score highly on comfort metrics.}.  Fine-tuning on this data provides our model with targeted feedback on distinct driving aspects, offering a more fine-grained and effective learning signal than the single, overall objective used in standard DPO.

Without bells and whistles, our model achieves a state-of-the-art 89.8 PDMS on the NAVSIM dataset, using the same backbone as competing methods. Comprehensive ablation studies validate the effectiveness of both the TISA module and the multi-objective post-training process in significantly advancing end-to-end planning capabilities.

Our contributions are three-fold:
\begin{itemize}
    \item We formalize and address the spatio-temporal misalignment problem in autoregressive end-to-end planning, a critical issue caused by conditioning future actions on a stale worldview.
    \item We propose a novel and efficient TISA module that performs view transformations in the latent space, significantly improving planning performance without notable computational overhead.
    \item We introduce a multi-objective DPO strategy that uses targeted, fine-grained preference pairs to refine the driving policy, demonstrably outperforming the standard single-objective DPO baseline.
\end{itemize}

\section{RELATED WORK}

In this section, we will first review early regression-based planning methods and their limitations. Then we will review the dominant probabilistic paradigms that embrace the inherent multi-modality of driving, including single-step, autoregressive, and diffusion-based approaches, thereby providing the context for our own contributions.

\subsection{Regression-based End-to-End Planning}

Early end-to-end planning methods framed the task as a regression problem, predicting a fixed set of future trajectories directly from historical sensor data. Initial works \cite{hu2023planning, jiang2023vad, shao2023reasonnet}, such as UniAD \cite{hu2023planning} and VAD \cite{jiang2023vad}, conditioned their predictions on dense BEV feature maps \cite{li2024bevformer, yang2023bevformer}. This approach suffers from high computational overhead due to redundant spatial information. Subsequent methods like SparseAD \cite{zhang2024sparsead} and SparseDrive \cite{sun2025sparsedrive} addressed this by first constructing sparse, instance-centric representations from the BEV features, focusing only on the most salient actors and map elements.

It is known that all regression-based methods share a fundamental limitation. By optimizing a regression loss (e.g., L2 distance), they are implicitly trained to predict the mean of possible future trajectories. In realistic scenarios where multiple distinct paths are valid (e.g., turning left or right), this leads to the well-known mode-averaging problem: the predicted average trajectory may be a dynamically infeasible or unsafe path that lies between the valid modes.

\subsection{Probabilistic End-to-End Planning}
\label{sec:related_work_probabilistic}

To overcome the limitations of regression, modern approaches model the planning task probabilistically, primarily following three paradigms: single-step methods, autoregressive models, and diffusion models.

\vspace{0.15em}

\noindent \textbf{Single-Step Methods.}\quad
These methods discretize the entire planning horizon into a finite set of complete trajectory candidates. Works like VADv2 \cite{chen2024vadv2} and Hydra-MDP \cite{li2024hydra} first create a ``trajectory dictionary" or codebook by clustering trajectories from the training dataset. The model then predicts a probability distribution over this codebook. While capable of multi-modal prediction, the quality and diversity of these pre-defined trajectories are inherently limited by the coverage of the training data, limited in generalization to novel scenarios.

\vspace{0.15em}

\noindent \textbf{Autoregressive Models.}\quad
Autoregressive models avoids discretizing trajectories in the entire planning horizon, instead generating trajectories sequentially, one step at a time. This allows for a much larger and more flexible action space.

In modular, two-stage pipelines, models like SMART \cite{wu2024smart}, MotionLM \cite{zhao2024kigras}, and others \cite{zhao2024kigras,zhao2025drope} leverage this step-wise nature to explicitly update their internal world state at each future step. By reasoning over an inferred future, they demonstrate strong performance in complex, interactive scenarios \cite{zhao2025autoregressive}.
This explicit world state update, however, is largely intractable for end-to-end models due to the complexity of predicting future raw sensor data \cite{chen2024drivinggpt}. Consequently, prominent autoregressive end-to-end planners, such as those based on Vision-Language-Action (VLA) models \cite{zhou2025autovla,hwang2024emma,li2025recogdrive}, typically condition the entire future plan on the initial sensor snapshot. While being a practical option, it creates the spatio-temporal misalignment problem that motivates our work.

\vspace{0.15em}

\noindent \textbf{Diffusion Models.}\quad
A third emerging direction employs diffusion models for end-to-end planning \cite{liao2025diffusiondrive,jiang2025diffvla,jiang2025diffvla}. These approaches treat planning as an iterative denoising process \cite{ho2020denoising, song2021denoising}, starting with a sample from clustered anchors and progressively refining it to match the learned data distribution. While these models have shown impressive results in generating realistic trajectories, they model the probability distribution implicitly. This can make them less amenable to certain post-training refinement techniques \cite{rafailov2023direct, schulman2017proximal, christiano2017deep}, like reinforcement learning, that rely on explicit action probabilities.

\begin{figure*}[!t]
    \centering
    \includegraphics[width=0.99\linewidth]{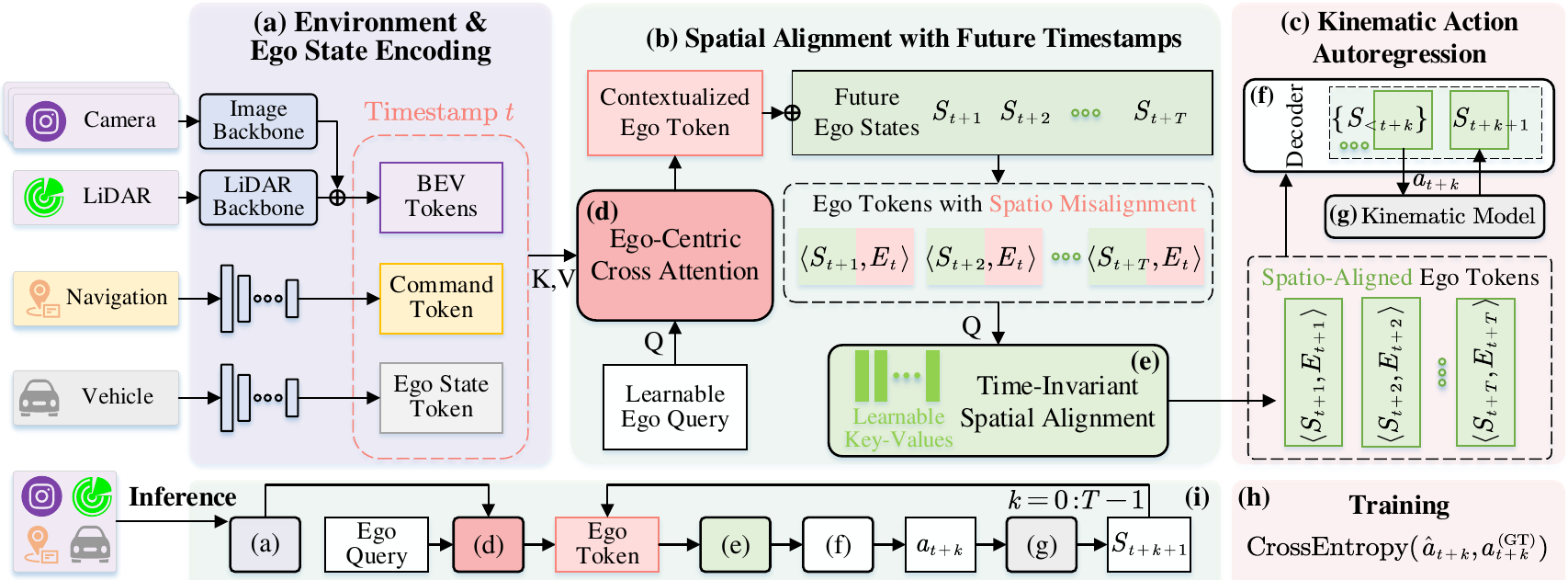}
    \caption{The diagram of our model architecture. \textbf{(a)} Sensor Data Encoding. \textbf{(b)} Spatial Alignment: For each future step $t+k$, the (predicted) ego state queries the TISA module (e) to receive an environmental context that is aligned to its future perspective, yet based on sensor data from time $t$.
    \textbf{(c)} Kinematic Action Prediction: The final decoder predicts probability distribution on the discretized action (acceleration, yaw rate). Their labels are inferred from the ground truth trajectories. \textbf{(d)} Ego-Centric Cross-Attention: A learnable query fuses sensor tokens to create a single Contextualized Ego Token representing the scene at time $t$. \textbf{(e)} TISA Key-Values: Shared, learnable Key-Value pairs that enable the TISA module to perform view alignment. \textbf{(f)} Transformer Decoder: Processes the sequence of view-aligned tokens to predict the action for each step. \textbf{(g)} Kinematic Model: Get the next ego state using the predicted action, enabling the autoregressive loop. \textbf{(h)} Training: The model is trained with cross-entropy loss on discretized actions derived from ground-truth trajectories. \textbf{(i)} Inference: The model generates a plan by autoregressively predicting an action and updating its state via the kinematic model. }
    \label{fig:framework}
\end{figure*}

\section{MODEL ILLUSTRATION}

In this section, we detail our proposed framework for autoregressive end-to-end planning. We begin by reformulating the planning problem to focus on predicting a sequence of kinematic actions, a formulation that ensures physically controllable trajectories. We then present our model architecture, centered around a novel Time-Invariant Spatial Alignment (TISA) module designed to resolve critical spatio-temporal misalignments. Finally, we introduce our Multi-Objective Post-Training stage, which employs Direct Preference Optimization (DPO) to refine the model's policy, instilling it with nuanced driving behaviors that surpass simple imitation.

\subsection{Task Formulation with Kinematic Action}

The goal of end-to-end planning is to predict the ego-vehicle's future trajectory given a history of sensor data. The input consists of multi-view camera images, LiDAR point clouds, a high-level navigation command, and the ego-vehicle's current state (e.g., heading $\theta$ and velocity $v$). At a given timestamp $t$, the model predicts a future trajectory as a sequence of states $\{s_{t+k}^{(t)} = (x_{t+k}^{(t)}, y_{t+k}^{(t)}, v_{t+k}^{(t)},\theta_{t+k}^{(t)})\}_{k=1}^{T}$. These predicted waypoints are \textit{expressed in the ego-vehicle's coordinate frame at the initial timestamp} $t$.

Instead of directly predicting the trajectory, we reframe it as predicting a sequence of kinematic actions. We model the ego-vehicle's motion with a kinematic model \cite{}, $\mathcal{K}$:
\begin{equation}
    (\Delta x, \Delta y, \Delta v, \Delta \theta) = \mathcal{K}(v, \Delta t; a, \dot{\psi}),
    \label{eq:kinematic_model}
\end{equation}
where $(\Delta x, \Delta y, \Delta v, \Delta \theta)$ represents the change in the ego-vehicle's state over a single time step of duration $\Delta t$, measured in its \textit{local coordinate frame at that instant}. This change is governed by the current velocity $v$ and the control inputs: longitudinal acceleration $a$ and yaw rate $\dot{\psi}$. We assume these control inputs are constant over each time step, forming a discrete action $A_k = (a_k, \dot{\psi}_k)$.

A sequence of these actions, $\{A_i\}_{i=t}^{t+T-1}$, can be unrolled using the kinematic model to reconstruct the full trajectory in the initial coordinate frame. This requires recursively applying Equation \eqref{eq:kinematic_model} and transforming the local position changes $(\Delta x, \Delta y)$ from each step back into the coordinate frame of the initial timestamp $t$ by applying rotation:
\begin{equation}
\begin{pmatrix} x_{i+1}^{(t)} \\ y_{i+1}^{(t)} \end{pmatrix} = \begin{pmatrix} x_i^{(t)} \\ y_i^{(t)} \end{pmatrix} + R\left(\sum_{j=t}^{i-1} \Delta \theta_j\right) \begin{pmatrix} \Delta x_i \\ \Delta y_i \end{pmatrix},
\label{eq:action_to_waypoint}
\end{equation}
where $(x_i^{(t)}, y_i^{(t)})$ is the vehicle's position at step $i$ expressed in the frame of time $t$, and $R(\cdot)$ is the 2D rotation matrix that accounts for the accumulated change in heading.

This formulation transforms the planning task into an autoregressive action prediction problem. The goal is to learn the conditional probability distribution of the next action, given the history of environments and ego-states:
\begin{equation}
   \prod_{k=0}^{T-1} P(A_{t+k} \mid E_{\le t+k}, s_{\le t+k}),
    \label{eq:task_definition}
\end{equation}
Critically, this step-by-step formulation requires a \textbf{causally consistent worldview} for making a decision at any future step $t+k$. This means that the conditioning information, especially the current observation of the environment and ego-state ($E_{t+k}, s_{t+k}$), should be represented within the ego-vehicle's reference frame at that specific instant, $t+k$.

\subsection{Architecture}
The architecture is presented in Fig. \ref{fig:framework} and detailed below.

\vspace{0.15em}

\noindent \textbf{Encoding.}\quad
Let the model's dimension be $d$. At the current time $t$, input sensor data is processed by backbones to produce a BEV token set $\text{Env}_t$, a high-level command token $\text{Cmd}_t$, and an ego-state token $S_t$. We employ a learnable ego query, $q_{\text{ego}} \in \mathbb{R}^d$, which uses cross-attention \cite{vaswani2017attention} to distill these varied inputs into a single Contextualized Ego Token, $\text{Ego}_t$, that summarizes the scene at the initial timestamp:
\begin{equation}
    \text{Ego}_t = \text{Attn}\left(q_{\text{ego}}, [\text{Env}_t, \text{Cmd}_t, S_t]\right),
    \label{eq:context_ego}
\end{equation}
where $[\cdot]$ denotes concatenation of the token sets.
To enrich the BEV token embeddings, we jointly train the model on two auxiliary tasks: object detection and map segmentation. This forces the BEV representation to learn explicit semantic information about the scene.

\vspace{0.15em}

\noindent \textbf{Time Invariant Spatial Alignment (TISA).}\quad
To plan for a future step $t+k$, we first form a prospective ego token by incorporating the predicted future state embedding, $S_{t+k} \in \mathbb{R}^d$, into the initial context:
\begin{equation}
    \text{Ego}_{t+k}^{(t)} = \text{Ego}_{t} + S_{t+k}^{(t)}.
\end{equation}
Crucially, the resulting token inherits its environmental understanding from $\text{Env}_t$. This creates the spatio-temporal misalignment: the token represents the ego-vehicle's state at a future time $t+k$ but perceives the world from a stale viewpoint at time $t$.

To fix this, we introduce the TISA module. It realigns the prospective ego token's worldview in the latent space. It consists of a set of $m$ learnable key-value pairs, $(K_{\text{TISA}}, V_{\text{TISA}}) \in \mathbb{R}^{m \times d}$, which act as a shared, learnable memory that the prospective token can query to perform the alignment:
\begin{equation}
    \text{Ego}_{t+k} = \text{Attn}(\text{Ego}_{t+k}^{(t)}, K_{\text{TISA}}, V_{\text{TISA}}).
\end{equation}
The intuition behind TISA is that the geometric transformation required to update an ego-centric view (a rotation and translation) is a time-invariant function of the relative pose change. Our prospective token $\text{Ego}_{t+k}^{(t)}$ already contains this necessary information, as it has fused the initial world context ($\text{Ego}_t$) with the future state embedding ($S_{t+k}^{(t)}$), which implicitly defines this pose change. Therefore, the TISA module learns a general, latent-space function that uses this information to perform the appropriate view transformation, yielding an aligned ego token, $\text{Ego}_{t+k}$, that perceives the environment from the correct future perspective.

\vspace{0.15em}

\noindent \textbf{Action Discretization and Label Derivation. }
We discretize the continuous action space of acceleration and yaw rate into a joint vocabulary of $m_a \times m_\omega$ actions. To generate training labels from the continuous ground-truth (GT) trajectories, we solve an inverse problem: for each step, we find the discrete action that, when rolled out for a 3-step horizon using our kinematic model $\mathcal{K}$, minimizes the L2 distance to the GT trajectory.
Finally, to ensure kinematic consistency, we replace the original GT trajectories with new ones generated from this derived action sequence. This guarantees that the model is trained on physically feasible trajectories that are perfectly reproducible by its own kinematic world model.

\vspace{0.15em}

\noindent \textbf{Decoder.}\quad
A Transformer decoder processes the sequence of aligned ego tokens $\{\text{Ego}_{t+k}\}_{k=0}^{T-1}$. At each step $k$, it outputs a probability distribution over the discrete action vocabulary.

\vspace{0.15em}

\noindent \textbf{Training.}\quad
The model is optimized via a cross-entropy loss between the predicted action distribution and the discrete action labels, $A_{t+k}^{(\text{GT})}$, derived from the GT trajectories.

\vspace{0.15em}

\noindent \textbf{Inference.}\quad
The process is autoregressive. At each step, an action $A_{t+k}$ is sampled from the predicted distribution. This action is then used by the kinematic model (Eq. \eqref{eq:action_to_waypoint}) to roll out the next ego-state $s_{t+k+1}$, which is then used to generate the aligned ego token for the next iteration of the loop.

\begin{table}[!t]
    \centering
    \caption{ Selection criteria for targeted "loser" trajectories in our multi-objective preference dataset.}
    \begin{tabular}{@{}ccccc@{}}
\toprule
Metrics & Collision & DA & EP     & TTC \\ \midrule
$y_{l,{\text{Coll}}}$  & 0    & 1  & 0      & 0   \\
$y_{l,{\text{DA}}}$    & 1    & 0  & 0      & 1   \\
$y_{l,{\text{EP}}}$    & 1    & 1  & 0      & 1   \\
$y_{l,{\text{TTC}}}$   & 1    & 1  & $\le$1 & 0   \\ \bottomrule
\end{tabular}
\flushleft \begin{small}
    The metrics include binary indicators for success (1) or failure (0) in Collision, Drivable Area (DA) violation, and Time-to-Collision (TTC), along with a continuous score for Ego Process (EP) success.
    Failures across the evaluation metrics can be correlated. For example, a collision event would also register as a failure in proactive safety metrics like Time-to-Collision.
\end{small}
    \label{tab:post_train_metrics}
\end{table}

\begin{table*}[!t]
\centering
\caption{Comparison results with state-of-the-art approaches on the NAVSIM benchmark. }
\label{tab:main_results}
\begin{tabular}{@{}l|ll|l|llllll@{}}
\toprule
Method               & Input  & Backbone      & Type        & NC   & DAC  & TTC  & Comf. & EP   & PDMS \\ \midrule
UniAD \cite{hu2023planning}                & C      & ResNet-34\cite{he2016deep}     & Regression           & 97.8 & 91.9 & 92.9 & \textbf{100}   & 78.8 & 83.4 \\
PARA-Drive\cite{weng2024drive}           & C      & ResNet-34\cite{he2016deep}     & Regression          & 97.9 & 92.4 & 93.0 & 99.8  & 79.3 & 84.0 \\
Transfuser\cite{chitta2022transfuser}           & C \& L & ResNet-34\cite{he2016deep}     & Regression          & 97.7 & 92.8 & 92.8 & \textbf{100}   & 79.2 & 84.0 \\
VADv2-$\mathcal{V}_{8192}$\cite{chen2024vadv2}          & C \& L & ResNet-34\cite{he2016deep}     & Score-based & 97.2 & 89.1 & 91.6 & \textbf{100}   & 76.0 & 80.9 \\
Hydra-MDP-$\mathcal{V}_{8192}$-W-EP\cite{li2024hydra} & C \& L & ResNet-34\cite{he2016deep}     & Score-based & 98.3 & 96.0 & 94.6 & \textbf{100}   & 78.7 & 86.5 \\
DiffusionDrive\cite{liao2025diffusiondrive}       & C \& L & ResNet-34\cite{he2016deep}     & Diffusion   & 98.2 & 96.2 & 94.7 & \textbf{100}   & 82.2 & 88.1 \\
GoalFlow\cite{xing2025goalflow}             & C \& L & ResNet-34\cite{he2016deep}     & Diffusion   & 98.3 & 93.8 & 94.3 & \textbf{100}   & 79.8 & 85.7 \\ 
ARTEMIS\cite{feng2025artemis}              & C \& L & ResNet-34\cite{he2016deep}     & Autoregressive          & 98.3 & 95.1 & 94.3 & \textbf{100}   & 81.4 & 87.0 \\
AutoVLA\cite{zhou2025autovla}              & C      & Qwen2.5-VL-3B\cite{bai2025qwen2} & Autoregressive          & \textbf{98.4} & 95.6 & \textbf{98.0} & 99.9  & 81.9 & 89.1 \\
DrivingGPT\cite{chen2024drivinggpt}           & C      & LlamaGen\cite{sun2024autoregressive}      & Autoregressive          & 98.2 & \textbf{97.8} & 95.2 & 99.8  & 83.5 & 89.6 \\
\textbf{Ours}                 & C \& L & ResNet-34\cite{he2016deep}     & Autoregressive          & \textbf{98.4} & 97.6 & 94.7 & \textbf{100}   & \textbf{84.6} & \textbf{89.8} \\ \bottomrule
\end{tabular}
\flushleft\begin{small}
    `C' denotes camera and `L' denotes LiDAR input. ``Score-based methods" refers to the single-step probabilistic models detailed in Section \ref{sec:related_work_probabilistic}.
\end{small}
\end{table*}

\subsection{Multi-Objective Post-Training}

For our multi-objective post-training, we construct a preference dataset with highly specific comparison pairs for each scene. After sampling 128 candidate trajectories, we create these pairs from multiple objectives in Table \ref{tab:post_train_metrics}.

The pairing process is as follows:
1.  \textbf{Create a Winner Pool ($\mathcal{Y}_w$):} We identify a pool of winners by selecting the top 5 trajectories with the best overall scores across all metrics.
2.  \textbf{Select Targeted Losers ($\mathcal{Y}_l$): }For each individual safety objective $m$, we select a corresponding ``loser" trajectory, $y_{l,m}$. Crucially, this loser is not catastrophically bad; it is chosen for being deficient only in partial specific metrics while remaining compliant with other metrics.

This strategy provides a highly targeted learning signal for DPO. For example, a pair $(y_w, y_{l, \text{TTC}})$ teaches the model to improve its safety buffer, given that other basic safety rules are already being followed. This allows us to fine-tune unique, meaningful aspects of the driving policy. We then apply the DPO loss over all such generated pairs:

\begin{small}
    \begin{equation}
\begin{aligned}
&L_{\text{DPO}} =\\
&-\mathbb{E}_{(x,\mathcal{Y}_w,\mathcal{Y}_l)\sim \mathcal{D}} \left[\log\, \sigma\, \left(
  \beta \left( \mathbb{E}_{y_w \in \mathcal{Y}_w}\left[\log \frac{\pi_\theta(y_w|x)}{\pi_{ref}(y_w|x)}\right] \right.\right.\right.\\
&\qquad\qquad\qquad\qquad\qquad  - \left.\left.\left. \mathbb{E}_{y_l \in \mathcal{Y}_l}\left[\log \frac{\pi_\theta(y_l|x)}{\pi_{ref}(y_l|x)}\right]\right) \right)\right],
\end{aligned}
\label{eq:dpo_group_loss}
\end{equation}
\end{small}
where $\mathcal{Y}_w$ is the winner pool, $\mathcal{Y}_l$ is the set of targeted losers for a given scene $x$, $\pi_\theta$ is the policy being optimized, $\pi_{ref}$ is the reference policy, and $\beta$ is a temperature parameter.

This objective function collectively pushes the policy to prefer the characteristics of the high-performing trajectories over the specific, isolated failure modes in the loser set.

\begin{table}[!t]
\centering
\caption{Ablation Results on Main Modules.}
\begin{tabular}{@{}llll|lllll@{}}
\toprule
OD        & MS        & TISA      & MOPT        & NC   & DAC  & TTC  & EP   & PDMS \\ \midrule
\ding{55} & \ding{55} & \ding{55} & \ding{55} & 97.6 & 93.4 & 93.4 & 79.5 & 85.0 \\ 
\ding{51} & \ding{55} & \ding{55} & \ding{55} & 97.4 & 92.3 & 92.3 & 78.0 & 83.3 \\
\ding{51} & \ding{51} & \ding{55} & \ding{55} & 97.4 & 93.2 & 93.0 & 78.6 & 84.1 \\ 
\ding{55} & \ding{51} & \ding{55} & \ding{55} & 97.8 & 94.4 & 93.3 & 80.1 & \underline{85.6} \\ \midrule
\ding{55} & \ding{51} & \ding{51} & \ding{55} & 98.0 & 95.5 & 93.8 & 81.1 & 86.8 \\
\ding{55} & \ding{51} & \ding{51} & \ding{51} & \textbf{98.4} & \textbf{97.6} & \textbf{94.7} & \textbf{84.8} & \textbf{89.8} \\ \bottomrule
\end{tabular}
\flushleft
\begin{small}
    The acronyms ``OD" and ``MS" refer to the object detection and map segmentation auxiliary tasks, respectively.
\end{small}
\label{tab:ablation}
\end{table}

\section{EXPERIMENTS}

To validate the efficacy of our proposed architecture and training strategy, we conducted a comprehensive set of experiments. Our evaluation is structured as follows: First, we describe the dataset, metrics, and implementation details that form the basis of our experiments. Second, we compare our final model against leading state-of-the-art methods on the NAVSIM benchmark to establish its overall performance. Third, we perform detailed ablation studies to isolate and quantify the contributions of our proposed TISA module and Multi-Objective Post-Training (MOPT). Fourth, we further analyze the training dynamics of our multi-objective DPO. Finally, we present qualitative results to visually demonstrate the improvements in the model's driving behavior.

\subsection{Dataset and Metrics}

We validate our method on the \textbf{NAVSIM} dataset \cite{dauner2024navsim}, a large-scale benchmark for autonomous driving planning. Derived from OpenScene, it features challenging, dynamic scenarios with rich multi-modal sensor data, including 360-degree camera views and fused LiDAR, sampled at 2Hz. The task is to predict a 4-second trajectory from 2 seconds of historical input. We use the official data split of 1,192 training and 136 testing scenarios.

For evaluation, we employ the official \textbf{Predictive Driving Model Score (PDMS)}, an efficient and effective proxy for closed-loop simulation. The PDMS is a composite score aggregating five sub-metrics: No at-fault Collisions (NC), Drivable Area Compliance (DAC), Time-to-Collision (TTC), Ego Progress (EP), and Comfort (Comf.). Together, these provide a holistic assessment of the planner's safety, efficiency, and feasibility.

\subsection{Implementation Details}

\vspace{0.15em}

\noindent \textbf{Input Processing. }\quad
 Input data consists of three forward-facing camera images, which are cropped, downscaled, and concatenated into a $1024 \times 256$ resolution image, and a rasterized BEV LiDAR map.

\vspace{0.15em}

\noindent \textbf{Backbone. }\quad
For a fair and rigorous comparison with prior methods, our model adopts the \textbf{ResNet-34} backbone to extract features from the camera images, a common lightweight backbone also used in other state-of-the-art methods.

\vspace{0.15em}

\noindent \textbf{TISA Setting. }\quad
Our TISA module is implemented with 16 learnable key-value pairs, whose parameters are shared across all prediction time steps.

\vspace{0.15em}

\noindent \textbf{Action Discretization. }\quad
We discretize the continuous action space into a joint vocabulary. Acceleration is uniformly discretized into 128 bins over the range $[-12.5, 12.5] \text{ m/s}^2$, and yaw rate into 64 bins over $[-1.5, 1.5] \text{ rad/s}$, creating $128 \times 64$ unique kinematic actions. Training data segments with actions exceeding these ranges are filtered out.

\vspace{0.15em}

\noindent \textbf{Action Label Inference. }\quad
To generate training labels of kinematic actions, we derive discrete actions from the continuous GT trajectories. For each step, we solve an inverse problem by selecting the discrete action that, when rolled out for a 3-step horizon via our kinematic model, minimizes the L2 distance to the GT path. To ensure perfect kinematic consistency for the training process, we then replace the original GT trajectories with these newly generated, physically reachable paths.

\vspace{0.15em}

\noindent \textbf{Auxiliary Task Setting. }\quad
We employ two auxiliary tasks to enrich the BEV representation. For object detection, dynamic agents are perceived via a query-based detection head with a set-based loss.
For map segmentation, a convolutional segmentation head with a cross-entropy loss parses static map elements (e.g., drivable area).

\vspace{0.15em}

\noindent \textbf{Optimizer. }\quad
We train our model using the AdamW \cite{loshchilov2019decoupled} optimizer with a weight decay of $10^{-2}$. The learning rate is $6 \times 10^{-4}$ for pre-training and is reduced to $6 \times 10^{-5}$ for DPO fine-tuning, governed by a cosine annealing scheduler with a 5-epoch warmup and a minimum learning rate of $10^{-6}$.

\vspace{0.15em}

\noindent \textbf{Training Environment. }\quad
All experiments were conducted on a cluster of 8 NVIDIA V100 or A800 GPUs. We used a per-GPU batch size of 32, resulting in an effective total batch size of 256.

\begin{figure*}[!t]
    \centering
    \begin{subfigure}[b]{0.45\textwidth}
        \centering
        \includegraphics[width=0.94\linewidth]{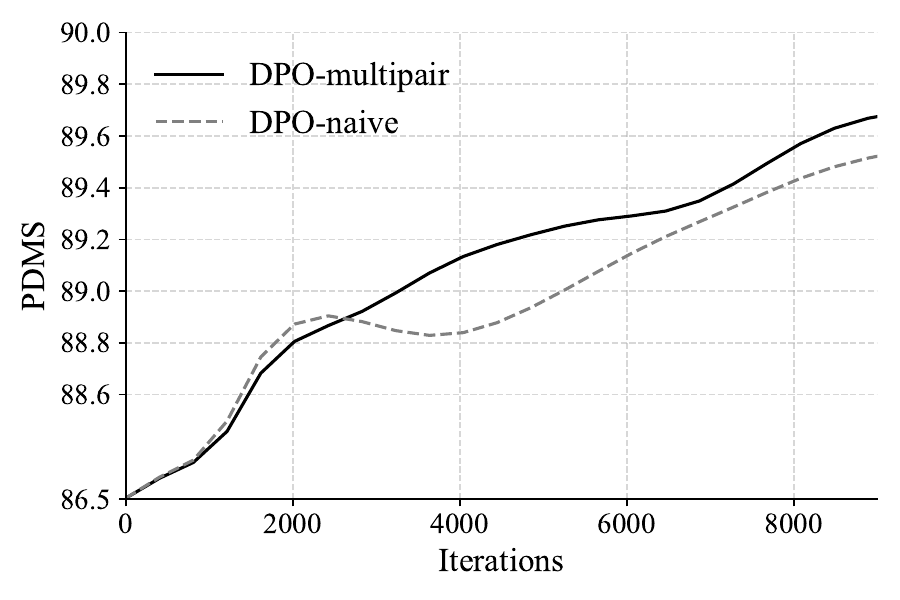}
        \caption{Comparison results on PDMS.}
        \label{fig:ablation_DPO_PDMS}
    \end{subfigure}
    \begin{subfigure}[b]{0.45\textwidth}
        \centering
        \includegraphics[width=0.98\linewidth]{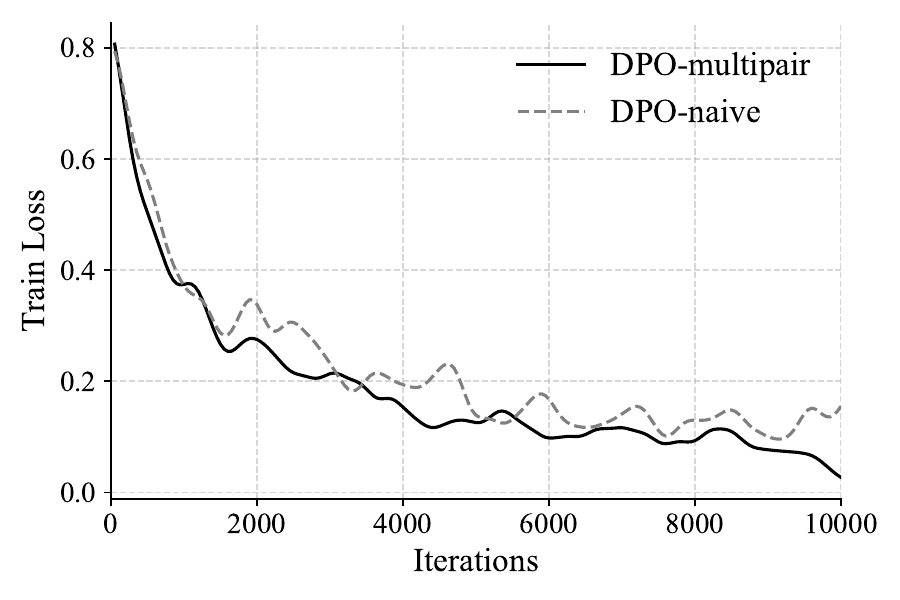}
        \caption{Comparison results on converging.}
        \label{fig:ablation_DPO_loss}
    \end{subfigure}
    \caption{Iterative comparison of our multi-objective DPO (DPO-multipair) and standard single-objective DPO (DPO-naive).}
    \label{fig:dpo_side_by_side}
\end{figure*}

\begin{figure*}[!t]
    \centering
    \includegraphics[width=0.95\linewidth]{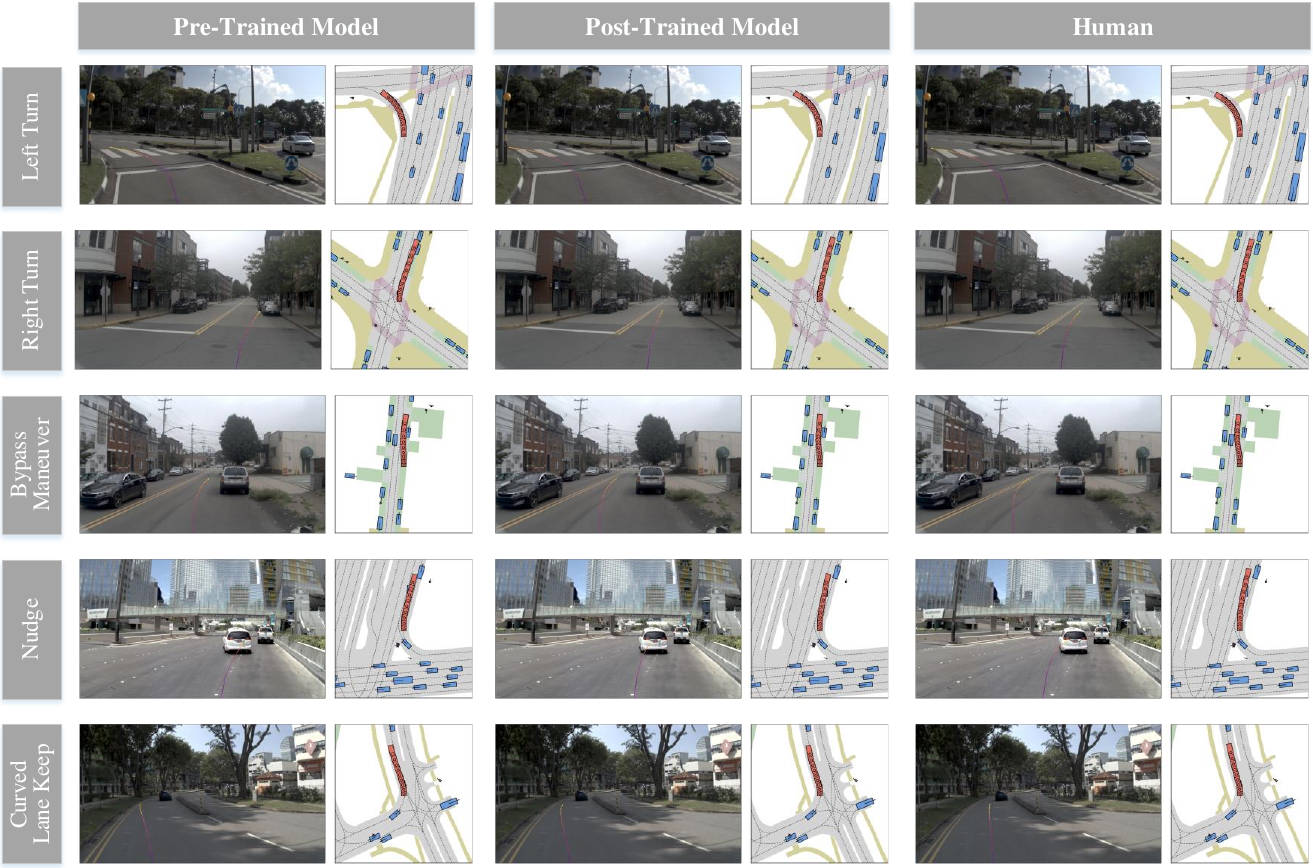}
    \caption{Qualitative results of various scenes between the pre-trained and post-trained model. The red rectangles in the BEV figure represent the future waypoints of the ego agent. Others represent agents in the current tiemstamp.}
    \label{fig:visualization}
\end{figure*}

\subsection{Comparison with State-of-the-Arts}

We benchmark our model against state-of-the-art approaches on the NAVSIM navtest split, encompassing regression, autoregressive (AR), and diffusion-based methods. To ensure a fair comparison, our model and the primary baselines utilize the same ResNet-34 backbone. Detailed results are presented in Table~\ref{tab:main_results}.

Our model establishes a new state-of-the-art for ResNet-34-based methods with a PDMS score of 89.8, significantly outperforming the next-best autoregressive model, ARTEMIS, by 2.8 points. More notably, our approach surpasses even models that leverage massive pre-trained Vision-Language Model (VLM) backbones, such as AutoVLA and DrivingGPT. This result underscores the efficacy of our architecture, which achieves top-tier performance without reliance on large-scale pre-training.

Furthermore, our model outperforms leading diffusion-based approaches like DiffusionDrive, demonstrating particular strengths on crucial DAC and EP metrics.

\subsection{Ablations on Various Modules}

We conduct an ablation study to validate the effectiveness of our primary contributions: the TISA module and our MOPT process. Our baseline model is first optimized by determining the best combination of auxiliary tasks (map segmentation and object detection), which we found to be using semantic segmentation alone. We then incrementally add our proposed modules to this strong baseline.

As shown in Table \ref{tab:ablation}, our optimized baseline achieves a PDMS of 85.6.
Integrating the TISA module boosts performance to 86.8 (+1.2), demonstrating its effectiveness in resolving spatio-temporal misalignment.
Subsequently, applying MOPT further elevates the score to our final result of 89.8 (+3.0).
The substantial gain from MOPT confirms the high potential of policy refinement for generative models. Notably, our TISA module provides a significant 1.2-point improvement from an architectural change alone, without requiring any additional fine-tuning. These results validate that both proposed components contribute significantly to the final model's state-of-the-art performance.

\subsection{Evaluation of Multi-Objective DPO}

To validate our multi-objective DPO strategy (hereafter DPO-multipair), we compare it against a standard single-pair DPO baseline (DPO-naive). We analyze both the final performance and the training dynamics, plotting the PDMS score over training iterations in Fig. (\ref{fig:ablation_DPO_PDMS}) and the training loss curves in Fig. (\ref{fig:ablation_DPO_loss}).

The results show that DPO-Multi consistently outperforms DPO-Single, achieving an approximately 0.2-point higher PDMS score at equivalent training stages. Furthermore, our approach exhibits more stable and rapid convergence, as indicated by the loss curves.
These findings highlight the benefit of our multi-objective method. By creating targeted preference pairs that isolate specific driving aspects (e.g., improving TTC while maintaining other aspects), DPO-multipair provides the model with fine-grained, disentangled learning signals. In contrast, the standard single-pair method offers only a coarse, overall ``good vs. bad" signal. This targeted feedback allows the model to explicitly refine distinct driving skills, leading to more stable optimization and a superior final policy.

\subsection{Qualitative Results}

Fig. \ref{fig:visualization} presents a qualitative comparison between our pre-trained model without MOPT and the final, post-trained model across five representative and challenging driving scenarios: a left turn from the dedicated left-turn lane, a right turn, a bypass maneuver, a nudge, and curved lane keeping.

Our analysis reveals two key improvements from the multi-objective post-training. First, the fine-tuned model remediates critical failures exhibited by the pre-trained model. In challenging scenarios like the right turn and the curved lane keeping, the pre-trained model produces trajectories that risk collision or road departure. The post-trained model consistently corrects these unsafe behaviors, executing the maneuvers robustly.
Second, beyond correcting outright failures, the final model demonstrates more nuanced, human-like driving. During the bypass and nudge maneuvers, the pre-trained model operates with little safety buffer. In contrast, our final model proactively maintains a large, safe margin from other agents, exhibiting more risk-averse and human-like behavior. This refinement is also evident in the smoother, more confident path tracking shown in the left turn from the dedicated left-turn lane.

\section{Discussions}

This section first discusses the limitations of our approach and suggests avenues for future research that build upon our contributions, and then concludes the paper.

\subsection{Limitations and Future Work}

While our method demonstrates strong performance, it has certain limitations. The efficacy of our multi-objective DPO strategy is dependent on the quality of pre-labeling outputs (e.g., object detection and map segmentation) used for labeling preference pairs. Errors in these perception modules can propagate, potentially leading to noisy preferences that could limit the impact of fine-tuning. Furthermore, while our TISA module effectively mitigates spatio-temporal misalignment, it does so in the latent space. The fundamental challenge, the absence of true future sensor data, persists, meaning the model cannot react to unpredictable events that are unobservable at the initial time step.

This leads to a promising direction for future research. The aligned latent space produced by TISA provides a geometrically consistent foundation at each future step. This foundation could be used not just for planning, but also as a basis for explicitly predicting future world states in a latent motion model. Such an approach could combine the benefits of our alignment technique with the proactive capabilities of a true world model, enabling the agent to reason about a wider range of future possibilities.

\subsection{Conclusions}
In this work, we addressed the critical challenge of spatio-temporal misalignment in autoregressive end-to-end planning. We first identified and formalized this issue, which arises from conditioning future plans on stale sensor data, and then proposed the Time-Invariant Spatial Alignment (TISA) module, an efficient latent-space mechanism that corrects the agent's worldview for future steps.
Furthermore, we enhanced the driving policy using a novel multi-objective Direct Preference Optimization (DPO) strategy. By leveraging targeted, fine-grained preference pairs, this method proved more effective at instilling nuanced and safe behaviors than standard single-objective fine-tuning. Our final model, combining these contributions, establishes a new state-of-the-art on the NAVSIM benchmark, demonstrating that our architectural design and structured policy refinement can achieve top-tier performance without bells and whistles.

\bibliographystyle{ieeetr}
\bibliography{ref}

\end{document}